\documentclass[sigconf,nonacm,natbib=false]{acmart}
\makeatletter
\global\@ACM@balancefalse
\makeatother

\usepackage{algorithm}
\usepackage{algpseudocode}
\usepackage{multirow}
\usepackage{array}
\usepackage{placeins}

\setcopyright{none}
\renewcommand\footnotetextcopyrightpermission[1]{}
\usepackage[
  backend=biber,
  style=numeric,
  sorting=none,
  giveninits=true,
  maxbibnames=3,
  minbibnames=3,
  doi=false,
  isbn=false,
  url=false,
  eprint=false
]{biblatex}
\DeclareNameAlias{author}{family-given}
\DeclareNameAlias{editor}{family-given}

\renewrobustcmd*{\bibinitperiod}{}
\renewrobustcmd*{\bibinitdelim}{}
\DeclareDelimFormat{multinamedelim}{\addcomma\space}
\DeclareDelimFormat{finalnamedelim}{\addcomma\space}
\DefineBibliographyStrings{english}{andothers={et\addabbrvspace al\adddot}}
\DeclareFieldFormat[article,inproceedings,incollection,thesis,misc]{title}{#1}
\DeclareFieldFormat{pages}{#1}

\AtBeginBibliography{\sloppy}

\newbibmacro*{compact:author-title}{%
  \printnames{author}%
  \setunit{\addperiod\space}%
  \printfield{title}}

\DeclareBibliographyDriver{article}{%
  \usebibmacro{bibindex}\usebibmacro{begentry}%
  \usebibmacro{compact:author-title}\newunit\newblock
  \printfield{journaltitle}\setunit{\addcomma\space}\printfield{year}%
  \iffieldundef{volume}{}{\setunit{\addcomma\space}\printfield{volume}%
    \iffieldundef{number}{}{\printtext{\mkbibparens{\printfield{number}}}}}%
  \iffieldundef{pages}{}{\setunit{\addcolon\space}\printfield{pages}}%
  \usebibmacro{finentry}}

\DeclareBibliographyDriver{inproceedings}{%
  \usebibmacro{bibindex}\usebibmacro{begentry}%
  \usebibmacro{compact:author-title}\newunit\newblock
  \printfield{booktitle}\setunit{\addcomma\space}\printfield{year}%
  \iffieldundef{pages}{}{\setunit{\addcolon\space}\printfield{pages}}%
  \usebibmacro{finentry}}

\DeclareBibliographyDriver{misc}{%
  \usebibmacro{bibindex}\usebibmacro{begentry}%
  \usebibmacro{compact:author-title}%
  \iffieldundef{howpublished}{}{\setunit{\addperiod\space}\printfield{howpublished}}%
  \iffieldundef{year}{}{\setunit{\addcomma\space}\printfield{year}}%
  \usebibmacro{finentry}}

\title{Federated Topic Model and Model Pruning Based on Variational Autoencoder}

\author{Chengjie Ma}
\affiliation{%
  \institution{Beijing Key Laboratory of Intelligent Communication Software and Multimedia, Beijing University of Posts and Telecommunications}
  \city{Beijing}
  \country{China}}

\author{Yawen Li}
\authornote{Corresponding author: warmly0716@126.com.}
\affiliation{%
  \institution{School of Economics and Management, Beijing University of Posts and Telecommunications}
  \city{Beijing}
  \country{China}}

\author{Meiyu Liang}
\affiliation{%
  \institution{Beijing Key Laboratory of Intelligent Communication Software and Multimedia, Beijing University of Posts and Telecommunications}
  \city{Beijing}
  \country{China}}

\author{Ang Li}
\affiliation{%
  \institution{Beijing Key Laboratory of Intelligent Communication Software and Multimedia, Beijing University of Posts and Telecommunications}
  \city{Beijing}
  \country{China}}

\begin{abstract}
Topic modeling has emerged as a valuable tool for discovering patterns and topics within large collections of documents. However, when cross-analysis involves multiple parties, data privacy becomes a critical concern. Federated topic modeling has been developed to address this issue, allowing multiple parties to jointly train models while protecting privacy. However, there are communication and performance challenges in the federated scenario. To solve these problems, this paper proposes a method to establish a federated topic model while ensuring the privacy of each node and uses neural network model pruning to accelerate the model. The client periodically sends cumulative neuron gradients and model weights to the server, and the server prunes the model. To address different requirements, two methods are proposed to determine the pruning rate. The first slowly prunes throughout training, which has limited acceleration during training but can ensure higher accuracy and significantly reduce inference time. The second quickly reaches the target pruning rate early in training and then continues training with a smaller model. This approach may lose more useful information but can complete training faster. Experimental results show that the proposed variational-autoencoder-based federated topic model pruning can greatly accelerate training while maintaining model performance.
\end{abstract}


\begin{document}
\maketitle

\section{Introduction}

Classical topic models, including Latent Dirichlet Allocation (LDA)~\cite{blei2003lda} and Probabilistic Latent Semantic Analysis (pLSA)~\cite{srinivasarao2022emailPlsa}, have been widely used to uncover latent themes in document collections. Subsequent work has extended topic-oriented analysis to heterogeneous semantic text mining~\cite{li2021heterogeneousTopic}, sentiment-spike explanation for evolving public events~\cite{li2025sentimentSpike}, and hashtag recommendation from multiple microblog features~\cite{kou2018hashtagRecommendation}. Dynamic-interest tracking further supports multi-view clustering of scholars whose research profiles change over time~\cite{li2023scholarClustering}, while interpretable machine-learning models help expose the evidence behind intelligent decisions~\cite{li2019interpretableDecision}. With the emergence of deep learning, neural topic models (NTMs) have gained popularity because they use neural networks to learn relationships between documents and topics, aiming for higher-quality topic representations.

In fields such as science, technology, and innovation document analysis, topic models are used to compare topics in funded projects from different institutions and identify research strengths. Related representation techniques include heterogeneous graph attention for short-text classification~\cite{hu2019heterogeneousGraphAttention} and deep modularity learning for community detection~\cite{yang2016deepModularity}. Scientific-publication representation can also combine semantic-similarity attention with hypergraph convolution~\cite{li2026semanticHypergraph}. For incomplete graphs, teacher--student distillation can recover information from missing features and structure~\cite{huo2023t2gnn}, whereas retrieval-oriented masked autoencoding strengthens textual representations~\cite{xiao2022retroMae}. Self-supervised graph co-training provides another way to couple complementary relational signals~\cite{xia2021graphCoTraining}, and filter-enhanced multilayer perceptrons illustrate an efficient alternative for modeling sequential information~\cite{zhou2022filterMlp}. Constructing topic models for multiple document collections is nevertheless challenging because of privacy constraints.

Federated learning (FL) is a distributed framework in which central servers coordinate learning across data owners without collecting their raw data. Stochastic quantization can reduce the size of federated updates~\cite{li2022stochasticQuantization}, and memory-aware graph processing addresses scalability at the local-computation level~\cite{shao2021memoryAware}. Communication-efficient reinforcement federated learning further combines dynamic client selection with adaptive gradient compression~\cite{pan2025rfcsc}. Broader distributed-learning foundations include consensus under switching topologies~\cite{meng2016consensusSeeking}, multi-view neural prediction~\cite{li2022fuelConsumption}, and learnable-edge collaborative filtering~\cite{xiao2022lecf}. At the graph level, federated cross-graph node classification~\cite{guan2021federatedGnn} and distributed multimodal path queries~\cite{li2022distributedPaths} demonstrate coordination across structurally different sources. Reinforcement-based active client selection has also been developed for federated heterogeneous graph learning~\cite{wang2025activeClientSelection}, while heterogeneous graph attention matching supports multilingual retrieval~\cite{huang2021hgamn}.

Practical FL must coordinate privacy guarantees and reliable node updates. Neural learning systems for distributed business computing~\cite{li2018businessComputing} and robust filtering under noisy observations~\cite{li2017tobitKalman} provide relevant foundations. Communication-efficient FL~\cite{konecny2017communicationEfficiency}, federated optimization~\cite{konecny2016federatedOptimization}, privacy-preserving deep learning~\cite{shokri2015privacyPreserving}, average consensus~\cite{lin2009averageConsensus}, and multiagent tracking~\cite{meng2013trackingAlgorithms} address complementary aspects of this coordination. FedSIN additionally learns information-network representations through federated self-adaptation~\cite{li2026fedSin}. Researchers have explored FL for topic modeling through non-negative matrix factorization~\cite{si2022federatedNmf}, federated LDA~\cite{wang2020federatedLda}, and a semantic-consistent generic topic model~\cite{shi2020federatedTopicDiscovery}.

Privacy-aware modeling increasingly intersects with multimodal information processing. Federated supervised cross-modal retrieval learns shared retrieval representations without centralizing modality-specific data~\cite{li2024federatedCrossModal}. Outside topic modeling, bi-projection fusion for omnidirectional image super-resolution shows how complementary visual projections can be integrated~\cite{wang2024omnidirectionalSr}; crowd-counting research likewise illustrates attribute learning from complex visual observations~\cite{wei2019crowdCounting}. Scientific and technological information retrieval must additionally resist semantic and media adversaries across modalities~\cite{li2022crossMediaAdversarial}. These developments motivate privacy-preserving topic models that remain efficient when documents and computation are distributed.

Implementing federated algorithms in practice nevertheless presents challenges. Each participant must send a complete model-parameter update to the server during every global round, causing communication overhead and increased training time. These challenges motivate pruning federated topic models to speed up training and inference, enable faster convergence, and reduce resource requirements.

The main contributions are as follows:

\begin{itemize}
  \item We propose a federated-learning topic-model approach that applies model pruning. The topic model is first federated, after which model-pruning techniques are applied.
  \item We introduce progressive pruning for federated topic-model training. Clients periodically send weights and accumulated neuron gradients to the server, which prunes the neural topic model using this information. Pruning reduces communication and computation overhead on clients.
  \item We propose two ways to determine the pruning rate. Slow pruning throughout training favors accuracy and reduced inference time, whereas rapid initial pruning accelerates training at the cost of potentially discarding useful information.
\end{itemize}

\section{Federated Topic Model and Model Pruning Based on Variational Autoencoder}

\subsection{Federated Topic Model Based on Variational Autoencoder}

We propose Prune-FedAVITM, a federated topic model with model pruning based on variational autoencoding~\cite{srivastava2017avitm}. The aim is to train topic models more efficiently under limited computational resources and network bandwidth. Algorithm~\ref{alg:federated-topic} outlines the training process.

\begin{algorithm}[H]
\caption{Federated Topic Model}
\label{alg:federated-topic}
\begin{algorithmic}[1]
\Require Client corpora $C$, vocabulary $V$, number of topics $K$
\Ensure Global topic-word matrix $\Phi$ and document-topic matrix $\Theta$
\Statex \textit{// Phase 1: Vocabulary consensus}
\ForAll{clients $N_i\in\mathcal{N}$ in parallel}
  \State $V_i\gets\Call{GetClientVocab}{N_i}$
\EndFor
\State $V\gets\Call{Aggregate}{\{V_i\}}$
\State Send $V$ to every client $N_i\in\mathcal{N}$
\Statex \textit{// Phase 2: Federated training}
\ForAll{clients $N_i$ in parallel}
  \State Obtain global vocabulary $V$
  \State Train a local AVITM model $W_i$ using the local corpus
  \State Send local model $W_i$ to the server
\EndFor
\State The server receives all local topic models $W_i$
\State Aggregate the models using FedAvg: $W\gets\Call{Agg}{\{W_i\}}$
\State Extract the topic-word matrix $\Phi$ and document-topic matrix $\Theta$
\State Send global model $W$ to every client
\end{algorithmic}
\end{algorithm}

During vocabulary consensus, the server waits for word inputs from all nodes and merges them into a common vocabulary. The vocabulary initializes a global model with weights $W^{(0)}$. Once all clients receive the shared vocabulary and initial global model, federated training begins. During training, the FedAvg algorithm is employed. For each data batch, the server waits for clients to send their locally trained neural models. The models are aggregated, and updated global parameters are sent back to all clients. This is repeated until the global model converges or reaches a predefined number of iterations.

\subsection{Progressive Pruning of the Federated Topic Model}

This section describes how pruning is incorporated into federated topic-model training. Magnitude pruning~\cite{han2016deepCompression} removes neurons according to the absolute value of their weights. However, small-weight neurons may remain important during training, especially in later stages. Our strategy therefore tracks cumulative gradients during local training to identify potentially significant neurons. Algorithm~\ref{alg:progressive-pruning} describes this process.

\begin{algorithm}[H]
\caption{Progressive Pruning}
\label{alg:progressive-pruning}
\begin{algorithmic}[1]
\Require Number of iterations $I$, initial pruning rate $T$
\Ensure Trained and pruned federated neural topic model
\For{$k=0,\ldots,I-1$}
  \State Initialize each client's accumulated neuron gradients: $Z_i\gets\emptyset$
  \ForAll{clients $i$ in parallel}
    \State Calculate neuron gradients on local data and record $Z_i$
    \State Update local model parameters $W_i$
    \If{this is a pruning round}
      \State Upload $W_i$ and $Z_i$ to the server; set $Z_i\gets\emptyset$
    \Else
      \State Upload only $W_i$ to the server
    \EndIf
  \EndFor
  \State Server receives all client model parameters
  \State $w(k)\gets\Call{Agg}{\{W_i^k\}}$
  \If{this is a pruning round}
    \State Calculate target pruning rate $T_k$ for this round
    \State Aggregate gradients $Z^k\gets\Call{Agg}{\{Z_i^k\}}$
    \State Prune according to neuron weights in $w(k)$
    \State Recover neurons according to $Z^k$, with larger values indicating recovery
    \State Apply the pruning mask: $w(k)\gets w(k)\otimes m(k)$
  \EndIf
  \State Send global model $w(k)$ to each client for a new FL round
\EndFor
\end{algorithmic}
\end{algorithm}

Progressive pruning is performed alongside standard FedAvg. Pruning occurs between rounds, with the interval set as a multiple of the iterations per round. Each pruning operation identifies a set of remaining model parameters, after which training continues with the pruned model and mask until the next pruning operation.

The progressive algorithm performs multiple pruning iterations to reach the target pruning rate. Two strategies set the target rate for individual pruning iterations. The first retains as much information as possible during training and allocates the target pruning rate evenly across the entire process. If the target rate is 50\%, then 25\% is reached at the halfway point. We call this strategy NormalPrune-FedAVITM. The final target is achieved when training finishes, and inference time can be reduced significantly.

The second strategy quickly reaches the target pruning rate early and then continues training with a smaller model. It may lose more useful information but completes training faster. We call this FastPrune-FedAVITM. It can substantially reduce training time, although excessive loss of useful information can affect final accuracy.

\section{Experiments}

\subsection{Experimental Setup}

\textbf{Time of one FL round.} Model pruning aims to improve training speed. Because communication time is difficult to measure directly in simulated FL experiments, we use an approximation. When the remaining parameters of model $W$ are aggregate functions, the approximate time of one FL round is
\begin{equation}
T(W):=c+\sum_{j\in W}t_j,
\end{equation}
where $c\geq0$ is a fixed system-overhead constant and $t_j>0$ is the time corresponding to the $j$th parameter component. This linear function is sufficient according to prior work~\cite{jiang2022modelPruning}.

\textbf{Datasets.} Experiments use the 20NewsGroup (20NG) and DBLP datasets. Text and label data are divided into training and test sets, and classification accuracy on the test set is the evaluation metric.

\textbf{Default settings.} The number of clients is 10, client-local iterations are 10, and batch size is 64. Because DBLP and 20NG differ in size and number of categories, the unpruned federated topic model is first trained for 400 and 2500 rounds, respectively. Experiments verify convergence under different pruning rates. Both pruning strategies are evaluated with target densities of 0.8, 0.6, 0.4, 0.2, 0.1, and 0.01.

\subsection{Experimental Results}

\textbf{Comparison of AVITM and FedAVITM(1.0).} We compare final accuracy, average topic diversity, and average topic coherence for centralized AVITM~\cite{srivastava2017avitm} and FedAVITM. In FL, local models can be aggregated only periodically, and non-independent homogeneous data distribution among clients can reduce accuracy. Nevertheless, FedAVITM approaches or exceeds centralized AVITM on DBLP and 20NG, demonstrating its effectiveness (Table~\ref{tab:avitm-comparison}).

\begin{table*}[t]
\caption{AVITM and FedAVITM(1.0) on the two datasets}
\label{tab:avitm-comparison}
\centering
\resizebox{\textwidth}{!}{%
\begin{tabular}{lrrrrrr}
\toprule
\multirow{2}{*}{Model} & \multicolumn{2}{c}{Accuracy} & \multicolumn{2}{c}{Topic coherence} & \multicolumn{2}{c}{Topic diversity}\\
\cmidrule(lr){2-3}\cmidrule(lr){4-5}\cmidrule(lr){6-7}
 & DBLP & 20NewsGroup & DBLP & 20NewsGroup & DBLP & 20NewsGroup\\
\midrule
AVITM & 67.32 & 35.96 & 0.0448 & 0.00510 & 0.665 & 0.855\\
FedAVITM(1.0) & 64.15 & 43.40 & 0.0558 & 0.0674 & 0.5215 & 0.897\\
\bottomrule
\end{tabular}}
\end{table*}

\begin{figure*}[t]
\centering
\begin{minipage}[t]{0.32\textwidth}\centering
\includegraphics[width=\linewidth]{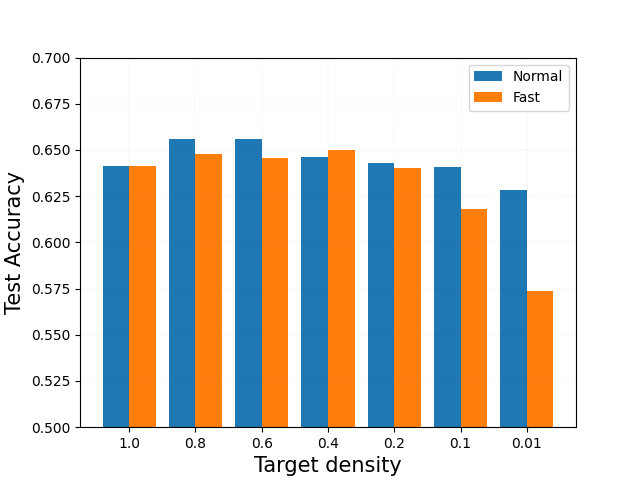}\\[-2pt](a) Test accuracy
\end{minipage}\hfill
\begin{minipage}[t]{0.32\textwidth}\centering
\includegraphics[width=\linewidth]{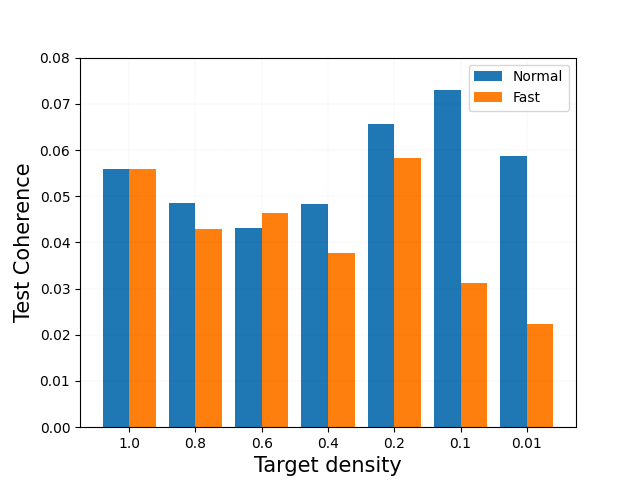}\\[-2pt](b) Topic coherence
\end{minipage}\hfill
\begin{minipage}[t]{0.32\textwidth}\centering
\includegraphics[width=\linewidth]{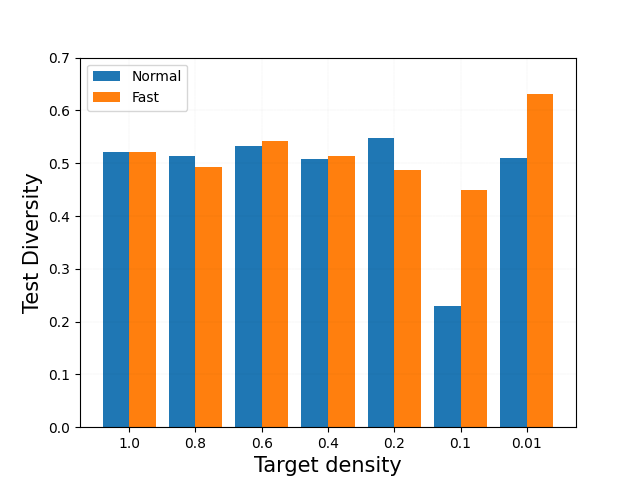}\\[-2pt](c) Topic diversity
\end{minipage}
\caption{Effect of different target densities on DBLP.}
\label{fig:dblp-density}
\Description{Three grouped bar charts report DBLP test accuracy, topic coherence, and topic diversity for normal and fast pruning at target densities from 1.0 to 0.01.}
\end{figure*}

\begin{figure*}[t]
\centering
\begin{minipage}[t]{0.32\textwidth}\centering
\includegraphics[width=\linewidth]{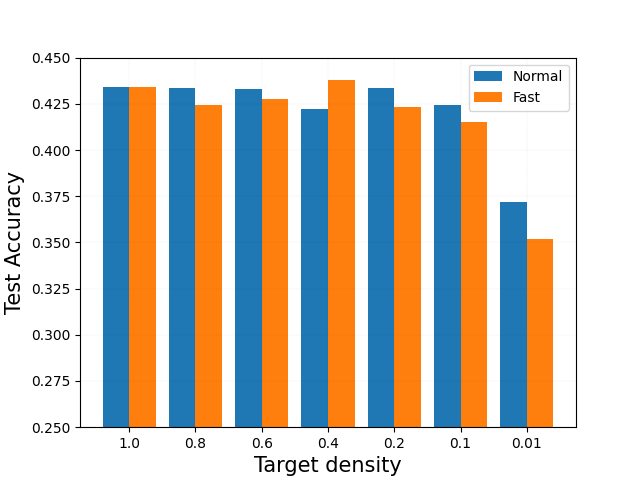}\\[-2pt](a) Test accuracy
\end{minipage}\hfill
\begin{minipage}[t]{0.32\textwidth}\centering
\includegraphics[width=\linewidth]{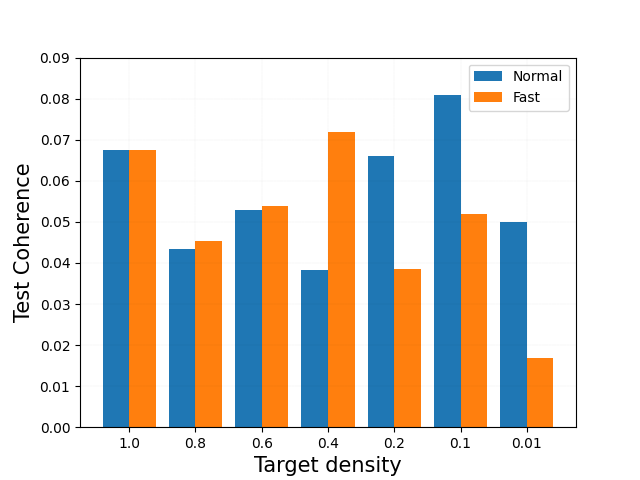}\\[-2pt](b) Topic coherence
\end{minipage}\hfill
\begin{minipage}[t]{0.32\textwidth}\centering
\includegraphics[width=\linewidth]{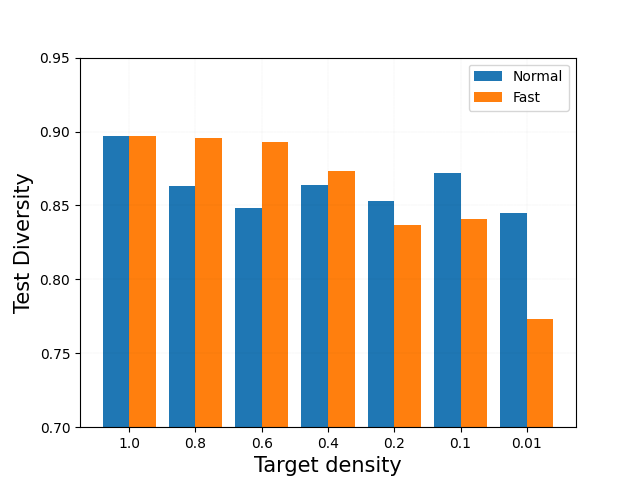}\\[-2pt](c) Topic diversity
\end{minipage}
\caption{Effect of different target densities on 20NewsGroup.}
\label{fig:20ng-density}
\Description{Three grouped bar charts report 20NewsGroup test accuracy, topic coherence, and topic diversity for normal and fast pruning at target densities from 1.0 to 0.01.}
\end{figure*}

\textbf{Best results at different target densities after model pruning.} Pruning speeds training and inference but may remove useful information and reduce final accuracy. Figures~\ref{fig:dblp-density} and~\ref{fig:20ng-density} show accuracy, topic coherence, and topic diversity at different target densities under the two pruning-speed strategies.

At larger target densities, accuracy can be better than without pruning. Accuracy gradually decreases as more neurons are removed. Prune-FedAVITM performs well when retaining 0.1 of the neurons, but performance drops sharply at 0.01. This suggests that training can recover some lost information and that removing less important neurons can improve the model. The normal method generally achieves higher accuracy, but the fast method performs better at a target density of 0.4, which may be an equilibrium point where sufficient early-stage information is retained.

The 20NG model shows a larger accuracy drop at density 0.1 than the DBLP model, perhaps because 20NG is more difficult and requires more retained neurons. Tables~\ref{tab:dblp-target-time} and~\ref{tab:20ng-target-time} report the time required to reach selected accuracy targets.

\begin{table*}[t]
\caption{Time to reach target accuracy on DBLP}
\label{tab:dblp-target-time}
\centering
\resizebox{\textwidth}{!}{%
\begin{tabular}{lrrrrrr}
\toprule
\multirow{2}{*}{Model} & \multicolumn{2}{c}{Time to 62\% accuracy} & \multicolumn{2}{c}{Time to 64\% accuracy} & \multicolumn{2}{c}{Time to 64.5\% accuracy}\\
\cmidrule(lr){2-3}\cmidrule(lr){4-5}\cmidrule(lr){6-7}
 & Normal & Fast & Normal & Fast & Normal & Fast\\
\midrule
FedAVITM(1.0) & \multicolumn{2}{c}{1077} & \multicolumn{2}{c}{2357} & \multicolumn{2}{c}{--}\\
Prune-FedAVITM(0.8) & 586 & 647 & 710 & 1062 & 1002 & 1879\\
Prune-FedAVITM(0.6) & 537 & 667 & 816 & 883 & 1646 & 2178\\
Prune-FedAVITM(0.4) & 615 & 575 & 1072 & 636 & 1951 & 745\\
Prune-FedAVITM(0.2) & 650 & 496 & 783 & 1125 & -- & --\\
Prune-FedAVITM(0.1) & 638 & -- & 900 & -- & 1672 & --\\
Prune-FedAVITM(0.01) & 609 & -- & 663 & -- & 716 & --\\
\bottomrule
\end{tabular}}
\end{table*}

\begin{table*}[t]
\caption{Time to reach target accuracy on 20NewsGroup}
\label{tab:20ng-target-time}
\centering
\resizebox{\textwidth}{!}{%
\begin{tabular}{lrrrrrr}
\toprule
\multirow{2}{*}{Model} & \multicolumn{2}{c}{Time to 40\% accuracy} & \multicolumn{2}{c}{Time to 42\% accuracy} & \multicolumn{2}{c}{Time to 43\% accuracy}\\
\cmidrule(lr){2-3}\cmidrule(lr){4-5}\cmidrule(lr){6-7}
 & Normal & Fast & Normal & Fast & Normal & Fast\\
\midrule
FedAVITM(1.0) & \multicolumn{2}{c}{3406} & \multicolumn{2}{c}{7543} & \multicolumn{2}{c}{15338}\\
Prune-FedAVITM(0.8) & 5801 & 4635 & 10629 & 7871 & 14242 & --\\
Prune-FedAVITM(0.6) & 3963 & 5271 & 7801 & 9534 & 13405 & --\\
Prune-FedAVITM(0.4) & 4986 & 3401 & 11150 & 5110 & -- & 8381\\
Prune-FedAVITM(0.2) & 3772 & 3524 & 7482 & 7483 & 10189 & --\\
Prune-FedAVITM(0.1) & 3614 & 5379 & 6102 & -- & 8863 & --\\
Prune-FedAVITM(0.01) & 3808 & -- & 6100 & -- & 7946 & --\\
\bottomrule
\end{tabular}}
\end{table*}

\textbf{Test accuracy versus time.} On DBLP, pruning generally outperforms unpruned FedAVITM unless target density is extremely low. On 20NG, accelerating training through pruning is more difficult, likely because the task is harder. Nevertheless, normal pruning with a low target density accelerates training before accuracy drops because of excessive pruning.

\begin{figure*}[t]
\centering
\begin{minipage}[t]{0.48\textwidth}\centering
\includegraphics[width=\linewidth]{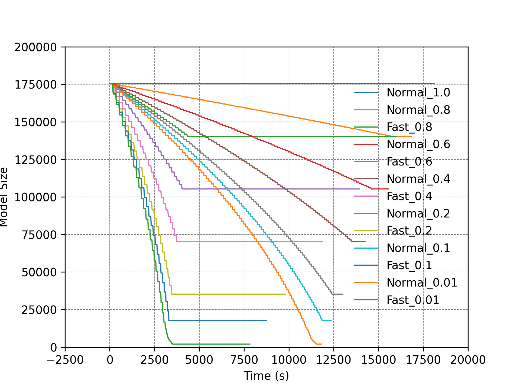}\\[-2pt](a) DBLP
\end{minipage}\hfill
\begin{minipage}[t]{0.48\textwidth}\centering
\includegraphics[width=\linewidth]{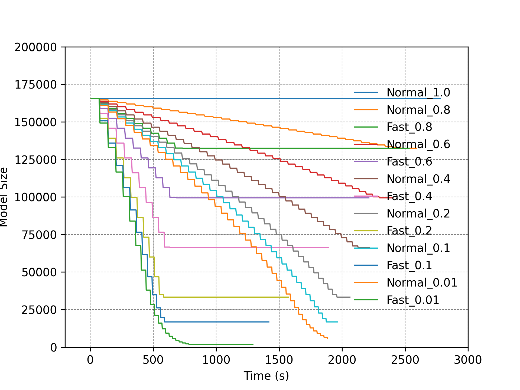}\\[-2pt](b) 20NewsGroup
\end{minipage}
\caption{Model size versus time for all target-density experiments.}
\label{fig:model-size-time}
\Description{Two line charts show model size over training time for normal and fast pruning configurations on DBLP and 20NewsGroup.}
\end{figure*}

\textbf{Model size and training time.} Figure~\ref{fig:model-size-time} shows parameter-count variation over time. Normal pruning generally achieves higher accuracy, whereas fast pruning reduces training time by completing pruning early and maintaining a smaller model. At target density 0.2, normal and fast training times are 72\% and 54\% of the unpruned time, respectively. Considering only data-access latency, the corresponding values are 60\% and 30\%. Table~\ref{tab:training-time} lists total simulation times.

\begin{table*}[t]
\caption{Time taken to complete training at different target densities}
\label{tab:training-time}
\centering
\small
\begin{tabular*}{\textwidth}{@{\extracolsep{\fill}}lrrrr@{}}
\toprule
\multirow{2}{*}{Model} & \multicolumn{2}{c}{20NewsGroup (2500 rounds)} & \multicolumn{2}{c}{DBLP (400 rounds)}\\
\cmidrule(lr){2-3}\cmidrule(lr){4-5}
 & Normal & Fast & Normal & Fast\\
\midrule
FedAVITM & \multicolumn{2}{c}{18109} & \multicolumn{2}{c}{2780}\\
Prune-FedAVITM(0.8) & 16934 & 16027 & 2592 & 2532\\
Prune-FedAVITM(0.6) & 15559 & 13946 & 2403 & 2212\\
Prune-FedAVITM(0.4) & 14462 & 11864 & 2215 & 1893\\
Prune-FedAVITM(0.2) & 13009 & 9782 & 2062 & 1574\\
Prune-FedAVITM(0.1) & 12491 & 8741 & 1963 & 1418\\
Prune-FedAVITM(0.01) & 11903 & 7819 & 1883 & 1294\\
\bottomrule
\end{tabular*}
\end{table*}

\FloatBarrier
\section{Conclusion}

This paper proposes a federated topic-model pruning method that applies traditional topic models to FL and uses neural-network pruning to construct and train federated neural topic models. Each node trains independently and periodically sends model parameters to the server, which uses them to prune the model and accelerate training. Two methods determine the pruning rate for different requirements. Slow pruning during training favors higher accuracy and reduced inference time, whereas rapid initial pruning accelerates training while potentially sacrificing useful information. Experiments show that the proposed method significantly accelerates training while maintaining performance, providing a new solution for applying federated learning to topic modeling.

\begin{acks}
This work was supported by the National Natural Science Foundation of China (62192784, U22B2038, and 62172056).
\end{acks}

\printbibliography[title={References}]

@article{li2023scholarClustering,
  author = {Li, Ang and Li, Yawen and Shao, Yingxia and Liu, Bingyan},
  title = {Multi-View Scholar Clustering with Dynamic Interest Tracking},
  journal = {IEEE Transactions on Knowledge and Data Engineering},
  year = {2023},
  volume = {35},
  number = {9},
  pages = {9671--9684}
}

@article{kou2018hashtagRecommendation,
  author = {Kou, Feifei and Du, Junping and Yang, Congxian and Shi, Yansong and Cui, Wanqiu and Liang, Meiyu and Geng, Yue},
  title = {Hashtag Recommendation Based on Multi-Features of Microblogs},
  journal = {Journal of Computer Science and Technology},
  year = {2018},
  volume = {33},
  pages = {711--726}
}

@article{li2021heterogeneousTopic,
  author = {Li, Yawen and Jiang, Di and Lian, Rongzhong and Wu, Xueyang and Tan, Conghui and Xu, Yi and Su, Zhiyang},
  title = {Heterogeneous Latent Topic Discovery for Semantic Text Mining},
  journal = {IEEE Transactions on Knowledge and Data Engineering},
  year = {2021},
  volume = {35},
  number = {1},
  pages = {533--544}
}

@article{blei2003lda,
  author = {Blei, David M. and Ng, Andrew Y. and Jordan, Michael I.},
  title = {Latent Dirichlet Allocation},
  journal = {Journal of Machine Learning Research},
  year = {2003},
  volume = {3},
  pages = {993--1022}
}

@article{srinivasarao2022emailPlsa,
  author = {Srinivasarao, U. and Sharaff, Aakanksha},
  title = {Email Thread Sentiment Sequence Identification Using {PLSA} Clustering Algorithm},
  journal = {Expert Systems with Applications},
  year = {2022},
  volume = {193},
  pages = {116475}
}

@article{li2022stochasticQuantization,
  author = {Li, Yawen and Li, Wenling and Xue, Zhe},
  title = {Federated Learning with Stochastic Quantization},
  journal = {International Journal of Intelligent Systems},
  year = {2022}
}

@article{shao2021memoryAware,
  author = {Shao, Yingxia and Huang, Shiyue and Li, Yawen and Miao, Xupeng and Cui, Bin and Chen, Lei},
  title = {Memory-Aware Framework for Fast and Scalable Second-Order Random Walk over Billion-Edge Natural Graphs},
  journal = {The VLDB Journal},
  year = {2021},
  volume = {30},
  number = {5},
  pages = {769--797}
}

@article{li2022fuelConsumption,
  author = {Li, Yawen and Zeng, Isabella Yunfei and Niu, Ziheng and Shi, Jiahao and Wang, Ziyang and Guan, Zeli},
  title = {Predicting Vehicle Fuel Consumption Based on Multi-View Deep Neural Network},
  journal = {Neurocomputing},
  year = {2022},
  volume = {502},
  pages = {140--147}
}

@article{xiao2022lecf,
  author = {Xiao, Shitao and Shao, Yingxia and Li, Yawen and Yin, Hongzhi and Shen, Yanyan and Cui, Bin},
  title = {{LECF}: Recommendation via Learnable Edge Collaborative Filtering},
  journal = {Science China Information Sciences},
  year = {2022},
  volume = {65},
  number = {1},
  pages = {1--15}
}

@inproceedings{guan2021federatedGnn,
  author = {Guan, Zeli and Li, Yawen and Xue, Zhe and Liu, Yuxin and Gao, Hongrui and Shao, Yingxia},
  title = {Federated Graph Neural Network for Cross-Graph Node Classification},
  booktitle = {IEEE 7th International Conference on Cloud Computing and Intelligent Systems},
  year = {2021},
  pages = {418--422}
}

@article{li2022distributedPaths,
  author = {Li, Yawen and Yuan, Ye and Wang, Yishu and Lian, Xiang and Ma, Yuliang and Wang, Guoren},
  title = {Distributed Multimodal Path Queries},
  journal = {IEEE Transactions on Knowledge and Data Engineering},
  year = {2022},
  volume = {34},
  number = {7},
  pages = {3196--3210}
}

@inproceedings{huang2021hgamn,
  author = {Huang, Jizhou and Wang, Haifeng and Sun, Yibo and Fan, Miao and Huang, Zhengjie and Yuan, Chunyuan and Li, Yawen},
  title = {{HGAMN}: Heterogeneous Graph Attention Matching Network for Multilingual {POI} Retrieval at Baidu Maps},
  booktitle = {27th ACM SIGKDD Conference on Knowledge Discovery and Data Mining},
  year = {2021},
  pages = {3032--3040}
}

@article{li2018businessComputing,
  author = {Li, Yawen and Jiang, Weifeng and Yang, Liu and Wu, Tian},
  title = {On Neural Networks and Learning Systems for Business Computing},
  journal = {Neurocomputing},
  year = {2018},
  volume = {275},
  number = {31},
  pages = {1150--1159}
}

@article{li2017tobitKalman,
  author = {Li, Wenling and Jia, Yingmin and Du, Junping},
  title = {Tobit Kalman Filter with Time-Correlated Multiplicative Measurement Noise},
  journal = {IET Control Theory and Applications},
  year = {2017},
  volume = {11},
  number = {1},
  pages = {122--128}
}

@misc{konecny2017communicationEfficiency,
  author = {Konečný, Jakub and McMahan, H. Brendan and Yu, Felix X. and Richtárik, Peter and Suresh, Ananda Theertha and Bacon, Dave},
  title = {Federated Learning: Strategies for Improving Communication Efficiency},
  howpublished = {arXiv:1610.05492},
  year = {2017}
}

@misc{konecny2016federatedOptimization,
  author = {Konečný, Jakub and McMahan, H. Brendan and Ramage, Daniel and Richtárik, Peter},
  title = {Federated Optimization: Distributed Machine Learning for On-Device Intelligence},
  howpublished = {arXiv:1610.02527},
  year = {2016}
}

@inproceedings{shokri2015privacyPreserving,
  author = {Shokri, Reza and Shmatikov, Vitaly},
  title = {Privacy-Preserving Deep Learning},
  booktitle = {22nd ACM SIGSAC Conference on Computer and Communications Security},
  year = {2015},
  pages = {1310--1321}
}

@inproceedings{lin2009averageConsensus,
  author = {Lin, Peng and Jia, Yingmin and Du, Junping and Yu, Fashan},
  title = {Average Consensus for Networks of Continuous-Time Agents with Delayed Information and Jointly-Connected Topologies},
  booktitle = {American Control Conference},
  year = {2009},
  pages = {3884--3889}
}

@article{meng2013trackingAlgorithms,
  author = {Meng, Deyuan and Jia, Yingmin and Du, Junping and Yu, Fashan},
  title = {Tracking Algorithms for Multiagent Systems},
  journal = {IEEE Transactions on Neural Networks and Learning Systems},
  year = {2013},
  volume = {24},
  number = {10},
  pages = {1660--1676}
}

@misc{li2022crossMediaAdversarial,
  author = {Li, Ang and Du, Junping and Kou, Feifei and Xue, Zhe and Xu, Xin and Xu, Mingying and Jiang, Yang},
  title = {Scientific and Technological Information Oriented Semantics-Adversarial and Media-Adversarial Cross-Media Retrieval},
  howpublished = {arXiv:2203.08615},
  year = {2022}
}

@article{wei2019crowdCounting,
  author = {Wei, X. and Du, J. and Liang, M. and Ye, Lingfei},
  title = {Boosting Deep Attribute Learning via Support Vector Regression for Fast Moving Crowd Counting},
  journal = {Pattern Recognition Letters},
  year = {2019},
  volume = {119},
  pages = {12--23}
}

@inproceedings{si2022federatedNmf,
  author = {Si, S. and Wang, J. and Zhang, R. and others},
  title = {Federated Non-Negative Matrix Factorization for Short Texts Topic Modeling with Mutual Information},
  booktitle = {International Joint Conference on Neural Networks},
  year = {2022},
  pages = {1--7}
}

@inproceedings{wang2020federatedLda,
  author = {Wang, Y. and Tong, Y. and Shi, D.},
  title = {Federated Latent Dirichlet Allocation: A Local Differential Privacy Based Framework},
  booktitle = {AAAI Conference on Artificial Intelligence},
  year = {2020},
  pages = {6283--6290}
}

@article{shi2020federatedTopicDiscovery,
  author = {Shi, Y. and Tong, Y. and Su, Z. and Jiang, D. and Zhou, Z. and Zhang, W.},
  title = {Federated Topic Discovery: A Semantic Consistent Approach},
  journal = {IEEE Intelligent Systems},
  year = {2020},
  volume = {36},
  number = {5},
  pages = {96--103}
}

@misc{srivastava2017avitm,
  author = {Srivastava, Akash and Sutton, Charles},
  title = {Autoencoding Variational Inference for Topic Models},
  howpublished = {arXiv:1703.01488},
  year = {2017}
}

@misc{han2016deepCompression,
  author = {Han, Song and Mao, Huizi and Dally, William J.},
  title = {Deep Compression: Compressing Deep Neural Networks with Pruning, Trained Quantization and Huffman Coding},
  howpublished = {arXiv:1510.00149},
  year = {2016}
}

@article{jiang2022modelPruning,
  author = {Jiang, Y. and Wang, S. and Valls, V. and Ko, B. J. and Lee, W. H. and Leung, K. K. and Tassiulas, L.},
  title = {Model Pruning Enables Efficient Federated Learning on Edge Devices},
  journal = {IEEE Transactions on Neural Networks and Learning Systems},
  year = {2022},
  pages = {1--13}
}

@article{meng2016consensusSeeking,
  author = {Meng, Deyuan and Jia, Yingmin and Du, Junping},
  title = {Consensus Seeking via Iterative Learning for Multi-Agent Systems with Switching Topologies and Communication Time-Delays},
  journal = {International Journal of Robust and Nonlinear Control},
  year = {2016},
  volume = {26},
  number = {17},
  pages = {3772--3790}
}

@article{li2026fedSin,
  author = {Li, Ang and Li, Yawen and Xue, Zhe},
  title = {{FedSIN}: Information Network Representation Based on Federated Self-Adaptive Learning},
  journal = {Frontiers of Computer Science},
  year = {2026},
  volume = {20},
  number = {1},
  pages = {2001307}
}

@article{li2024federatedCrossModal,
  author = {Li, Ang and Li, Yawen and Shao, Yingxia},
  title = {Federated Learning for Supervised Cross-Modal Retrieval},
  journal = {World Wide Web},
  year = {2024},
  volume = {27},
  number = {4},
  pages = {41}
}

@article{li2026semanticHypergraph,
  author = {Li, Ang and Li, Yawen and Kou, Feifei and Xue, Zhe and Liang, Meiyu and Wang, Baoxiang},
  title = {Semantic-Similarity Attention Meets Hypergraph Convolution for Scientific Publication Representation Learning},
  journal = {Frontiers of Computer Science},
  year = {2026}
}

@inproceedings{hu2019heterogeneousGraphAttention,
  author = {Hu, Linmei and Yang, Tianchi and Shi, Chuan and Ji, Houye and Li, Xiaoli},
  title = {Heterogeneous Graph Attention Networks for Semi-Supervised Short Text Classification},
  booktitle = {Conference on Empirical Methods in Natural Language Processing},
  year = {2019},
  pages = {4821--4830}
}

@inproceedings{zhou2022filterMlp,
  author = {Zhou, Kun and Yu, Hui and Zhao, Wayne Xin and Wen, Jirong},
  title = {Filter-Enhanced {MLP} Is All You Need for Sequential Recommendation},
  booktitle = {ACM Web Conference},
  year = {2022},
  pages = {2388--2399}
}

@inproceedings{yang2016deepModularity,
  author = {Yang, Liang and Cao, Xiaochun and He, Dongxiao and Wang, Chuan and Wang, Xiao and Zhang, Weixiong},
  title = {Modularity Based Community Detection with Deep Learning},
  booktitle = {International Joint Conference on Artificial Intelligence},
  year = {2016},
  pages = {2252--2258}
}

@inproceedings{xia2021graphCoTraining,
  author = {Xia, Xin and Yin, Hongzhi and Yu, Junliang and Shao, Yingxia and Cui, Lizhen},
  title = {Self-Supervised Graph Co-Training for Session-Based Recommendation},
  booktitle = {30th ACM International Conference on Information and Knowledge Management},
  year = {2021},
  pages = {2180--2190}
}

@inproceedings{huo2023t2gnn,
  author = {Huo, Cuiying and Jin, Di and Li, Yawen and He, Dongxiao and Yang, Yubin and Wu, Lingfei},
  title = {{T2-GNN}: Graph Neural Networks for Graphs with Incomplete Features and Structure via Teacher--Student Distillation},
  booktitle = {AAAI Conference on Artificial Intelligence},
  year = {2023},
  pages = {4339--4346}
}

@article{li2019interpretableDecision,
  author = {Li, Yawen and Yang, Liu and Yang, Bohan and Wang, Ning and Wu, Tian},
  title = {Application of Interpretable Machine Learning Models for the Intelligent Decision},
  journal = {Neurocomputing},
  year = {2019},
  volume = {333},
  pages = {273--283}
}

@inproceedings{xiao2022retroMae,
  author = {Xiao, Shitao and Liu, Zheng and Shao, Yingxia and Cao, Zhao},
  title = {{RetroMAE}: Pre-Training Retrieval-Oriented Language Models via Masked Auto-Encoder},
  booktitle = {Conference on Empirical Methods in Natural Language Processing},
  year = {2022},
  pages = {538--548}
}

@inproceedings{wang2024omnidirectionalSr,
  author = {Wang, Jiangang and Cui, Yuning and Li, Yawen and Ren, Wenqi and Cao, Xiaochun},
  title = {Omnidirectional Image Super-Resolution via Bi-Projection Fusion},
  booktitle = {AAAI Conference on Artificial Intelligence},
  year = {2024},
  volume = {38},
  number = {6},
  pages = {5454--5462}
}

@article{pan2025rfcsc,
  author = {Pan, Zhenhui and Li, Yawen and Guan, Zeli and Liang, Meiyu and Li, Ang and Wang, Jia and Kou, Feifei},
  title = {{RFCSC}: Communication Efficient Reinforcement Federated Learning with Dynamic Client Selection and Adaptive Gradient Compression},
  journal = {Neurocomputing},
  year = {2025},
  volume = {612},
  pages = {128672}
}

@inproceedings{wang2025activeClientSelection,
  author = {Wang, Jia and Li, Yawen and Shao, Yingxia and Xue, Zhe and Guan, Zeli and Li, Ang and Ye, Guanhua},
  title = {Reinforcement Active Client Selection for Federated Heterogeneous Graph Learning},
  booktitle = {AAAI Conference on Artificial Intelligence},
  year = {2025},
  volume = {39},
  number = {20},
  pages = {21117--21125}
}

@article{li2025sentimentSpike,
  author = {Li, Yawen and Wang, Xiaobao and Wen, Bin and Jin, Di and Du, Junping},
  title = {Sentiment Variation-Aware Sentiment Spike Explanation during {COVID-19} Epidemic},
  journal = {IEEE Transactions on Knowledge and Data Engineering},
  year = {2025},
  volume = {38},
  number = {2},
  pages = {1306--1318}
}
\end{document}